%% file: main.tex
\def\year{2022}\relax
\documentclass[letterpaper]{article} 
\usepackage{aaai22}  
\usepackage{times}  
\usepackage{helvet}  
\usepackage{courier}  
\usepackage[hyphens]{url}  
\usepackage{graphicx} 
\def\UrlFont{\rm}  
\usepackage{natbib} 
\usepackage{caption} 
\DeclareCaptionStyle{ruled}{labelfont=normalfont,labelsep=colon,strut=off}
\usepackage{algorithm} % Check
\usepackage{algpseudocode} % Check 
\usepackage[utf8]{inputenc}
\usepackage{appendix}
\usepackage{subfiles}
\usepackage{newfloat}
\usepackage{listings}
\usepackage{booktabs}
 \usepackage{multirow}
\usepackage{bbm}
\usepackage{bm}
\usepackage{amsfonts}
\usepackage{amsmath}
\usepackage{todonotes} 
\usepackage{tablefootnote}
\usepackage{xspace}
\usepackage{tabularx,colortbl}

\usepackage[utf8]{inputenc}
\usepackage[T1]{fontenc}
\usepackage{array, booktabs, ragged2e}
\floatstyle{ruled}
\newfloat{listing}{tb}{lst}{}
\floatname{listing}{Listing}
\newcommand{\PaperTitle}{Scope of Pre-trained Language Models for Detecting Conflicting Health Information}
\title{\PaperTitle}
\author{
    Joseph Gatto, Madhusudan Basak, Sarah M. Preum
}
\affiliations{
    Department of Computer Science, Dartmouth College
    
    \{joseph.m.gatto.gr, madhusudan.basak.gr, sarah.masud.preum\}@dartmouth.edu, 
}

\begin{document}
\maketitle

\begin{abstract}
\subfile{sections/Abstract}
\end{abstract}
\section{Introduction}

\subfile{sections/0Introduction}

% 

\begin{table}
\centering 
\begin{tabular}{lcc}
\toprule
\textbf{Dataset} & \textbf{Train} & \textbf{Test} \\
\midrule
Real &    1398 & 608  \\
Synthetic & 1427 & 470 \\
% $\mathrm{Test}_{Real}$   & 608 \\
% $\mathrm{Test}_{Synth}$  & 470 \\
\bottomrule 
\end{tabular}
\caption{Number of samples in each split of the HCD dataset. } 
\label{data_disribution}
\end{table}

\section{Problem Formulation}
\subfile{sections/1ProblemFormulation}

\section{Dataset}
\subfile{sections/3Dataset}

\section{Methods}

\subfile{sections/4Methods}
% 
% 

\section{Results}
\subfile{sections/5Results}

\section{Related Work}
\subfile{sections/2RelatedWork}

\begin{table*}[htbp]
\renewcommand{\arraystretch}{1.2}
\begin{tabularx}{\linewidth}{ X X l l}

\toprule
\textbf{Medical Professional A}  & \textbf{Medical Professional B} & \textbf{Post Topic} & \textbf{Conflict Type}\\ 
\midrule
For clarity's sake, are you saying you took 12x500 mg over the course of 23 hours or 11 hours? ... If it's the latter it may be worth getting checked as it's double the recommended dose. & Assuming the doses were spaced out evenly then concentrations would not have reached toxic levels at any point (see calculation below if you're interested). & Fever Medication & Direct \\
If the left breast has been getting larger and developing those veins recently, and if you feel firmness of the left breast that is different from the right, then I would get it checked out by your PCP or GYN.  & They’re normal. Breasts (like eyebrows) are “sisters, not twins” - no one has perfectly symmetrical breasts. & Breast Cancer & Conditional \\
\bottomrule
\end{tabularx}
\caption{Conflicting health information found on reddit.com/r/AskDocs --- A social media forum where anyone can get medical advice from a professional whose medical certifications are verified by the forum moderators. We show that suggestions from medical professionals can be conflicting. }  
\label{social_media_data}
\end{table*}

\section{Limitations and Future Work}
\subfile{sections/6Limitations_FutureWork}
% 

\section{Broader Impacts and Ethical Considerations}
\subfile{sections/7BroaderImpacts}
\section{Conclusion}

\subfile{sections/8Conclusion}
\bibliography{references}
% \newpage
% \appendix
% \subfile{sections/9Appendix}
\end{document}

%% file: sections/Abstract.tex
\begin{quote}

% Identifying conflicting medical information online is a safety-critical task for users with complex health backgrounds. In effort towards facilitating safer interactions with online health platforms, we  introduce a new task, Health Conflict Detection (HCD). Given two pieces of health advice, HCD asks 1) Is there a conflict present? and 2) What type of conflict occurs? The goal is to develop a system that can notify users of potentially harmful conflicts between two health suggestions. 

An increasing number of people now rely on online platforms to meet their health information needs. Thus identifying inconsistent or conflicting textual health information has become a safety-critical task. Health advice data poses a unique challenge where information that is accurate in the context of one diagnosis can be conflicting in the context of another. For example, people suffering from diabetes and hypertension often receive conflicting health advice on diet. This motivates the need for technologies which can provide \textbf{contextualized, user-specific health advice}. A crucial step towards contextualized advice is the ability to compare health advice statements and detect \textit{if and how they are conflicting}. This is the task of health conflict detection (HCD). Given two pieces of health advice, the goal of HCD is to detect and categorize the type of conflict. It is a challenging task, as (i) automatically identifying and categorizing conflicts requires a deeper understanding of the semantics of the text, and (ii) the amount of available data is quite limited. 

In this study, we are the first to explore HCD in the context of pre-trained language models. We find that DeBERTa-v3 performs best with a mean F1 score of 0.68 across all experiments. We additionally investigate the challenges posed by different conflict types and how synthetic data improves a model's understanding of conflict-specific semantics.  Finally, we highlight the difficulty in collecting real health conflicts and propose a human-in-the-loop synthetic data augmentation approach to expand existing HCD datasets. Our HCD training dataset is over 2x bigger than the existing HCD dataset and is made publicly available on Github.

% In this study, we explore four pre-trained language models' ability to detect conflicting health information. 

% we formulate health conflict detection as a multilabel multiclass classification task in the context of pre-trained language models. We explore the applicability of a state-of-the-art language model, namely BERT, to solve this problem. We introduce the Hierarchical Pair-Wise Encoder for Text Classification (\ModelName), which augments BERT by utilizing Abstract Meaning Representation (AMR) graphs for enhanced conflict detection. \ModelName shows promising results for such a resource-constrained text classification task and outperforms BERT by a (TODO)\% increase in F1 score on average. We also demonstrate how we can approach BERT-level performance on our pairwise inference task using only information extracted from AMR graphs

\end{quote}

%% file: sections/0Introduction.tex
% medinfo and inconsistent med info: misinformation, conflicting information --> HCD
In recent years, quick and easy access to online health information has changed the way people eat, exercise, and interact with medical professionals \cite{BujnowskaFedak2020}. As we continue to live through a global pandemic, people are increasingly relying on online resources for health advice \cite{Loomba2021}. However, a significant limitation of online health platforms is their inability to provide consistent, contextual health advice.  For instance, while many information sources promote taking daily aspirin after a heart attack, this advice is dangerous to those on warfarin, as combining the medicines can have life-threatening side-effects (See example 5 in Table \ref{dataset}). Moreover, as new clinical evidence emerges, some health advice will inevitably become outdated, generating conflicts with recent findings.

% Health advice provided online is most often without consideration of the end-users medical history. For instance, a user is advised to eat foods that interact poorly with their medications or trigger their food intolerance or allergy. 
% Alternatively, a user is advised to perform vigorous exercise that might be risky for them due to a pre-existing heart condition. 
Such situations motivate the need for automatic detection of conflicting health information across multiple relevant sources to facilitate safer interactions with online health platforms. This task is referred to as \textbf{health conflict detection} (HCD). 

Given two pieces of health advice statements, the goal of HCD is to both identify the presence of a conflict \textit{and} to recognize what type of conflict occurs \footnote{HCD does not aim at discounting a source of information as health conflicts can occur between a pair of advice even when both advice statements are correct \cite{carpenter2016conflicting}. Also, conflict resolution is beyond the scope of HCD as it requires the expertise of a professional healthcare provider and is often subjective to an individual.}. Different types of conflicts are presented in Table \ref{dataset}. Recognizing the conflict type will aid the user in interpreting the reason for a conflict correctly. For example, consider sample 5 in Table \ref{dataset}. Simply expressing that a conflict is present does not guide the user on how to react to the conflicting information on aspirin consumption. Explicitly stating the conflict type communicates how one should be careful regarding \textit{when} (temporal) aspirin should be taken as well as \textit{if} (conditional) aspirin should be consumed. With more people relying on online health websites and the surge of a global infodemic \cite{Buchanan2020}, detecting inconsistent/conflicting health information at large is even more critical now, as widespread health misinformation is a risk to public health and extremely expensive to combat \cite{cornish2020}. 
% In fact, in 2020 over \$471 million dollars were spent funding anti-misinformation campaigns worldwide. 

% why automatic HCD is important
% Conflicting health information may arise for a number of reasons, including but not limited to (i) the information sources pertaining to different health topics, (ii) one topic with conflicting underlying research, and (iii) misinformation where a information source may simply report false or conspiratory claims about a certain topic. The latter issue feels particularly relevant as the Covid-19 pandemic has also sparked an infodemic \cite{Buchanan2020}. Widespread health misinformation is a risk to public health and extremely expensive to combat. In fact, in 2020 over \$471 million dollars were spent funding anti-misinformation campaigns worldwide \cite{cornish2020}. 

\begin{table*}[htbp]
\renewcommand{\arraystretch}{1.1}
\begin{tabularx}{\linewidth}{l X X l}

\toprule
& \textbf{Advice 1}  & \textbf{Advice 2} & \textbf{Conflict Type}\\ 
\midrule
1 & Limit intake of sweetened drinks, \textbf{snacks} and desserts by eating them less often and in smaller amounts. & Many fruits need little preparation to become a healthy part of a meal or \textbf{snack}. & Direct \\
2 & Limit \textbf{liquids} before bed. & To prevent dehydration, drink plenty of \textbf{fluids} unless your doctor directs you otherwise. & Temporal \\ 
3 &Pasteurized dairy products, \textbf{milk}, cheese, yogurt, spinach are rich in calcium &
 Dilute full-fat \textbf{milk} in your cereal with water. This reduces the absorption of sugar and decreases fat intake. & Sub-typical \\

4 &  Eat dark green vegetables such as spinach, \textbf{collard greens, kale and broccoli} & Avoid sudden increase of \textbf{cruciferous vegetables} if you are on Coumadin. They may affect your treatment and dosage&
Conditional \\

%  Avoid excessive butter and \textbf{cheese}. &
% Eat calcium-rich foods like milk, \textbf{cheese} and green leafy vegetables.&
% Quantitative \\
5& \textbf{Aspirin} helps prevent future blood clots and decreases the risk of death after a heart attack. & \textbf{Aspirin}, aspirin-like drugs (salicylates), and nonsteroidal anti-inflammatory drugs ... may have effects similar to warfarin. These drugs may increase the risk of bleeding problems if taken during treatment with warfarin. & Conditional, Temporal\\
\bottomrule

\end{tabularx}
\caption{Example health conflict pairs from the HCD dataset. Each sample contains a pair of health advice regarding a common topic. The conflict label is annotated with respect to the common topic between pieces of advice. Conflict topics are bold in the examples above.}  
\label{dataset}
\end{table*}
%  challenges of automatic HCD in the context of PLM
% 

Conflicting health information appears in many forms across the internet. HCD was first introduced in \cite{7917875, PREUM2017226} where the authors introduce a novel HCD dataset using data extracted from several health apps and medical information websites. More recently, the need for conflict detection on other sources, including social media, has presented itself. Consider the sub-reddit r/AskDocs, a forum with 380k+ members where anyone can ask questions to a verified medical professional for free. In Table \ref{social_media_data} we highlight conflicting pairs of advice statements from registered nurses and licensed physicians identified on r/AskDocs to illustrate the need for more reliable and consistent social interaction on health. 

%  \cite{7917875, PREUM2017226}

In this study, we build on top of the initial HCD work proposed in \cite{7917875, PREUM2017226} by focusing on a scaleable approach to HCD data expansion. \cite{7917875, PREUM2017226} use a rule based solution to detect health conflicts. Although the proposed solution results in high recall the following issues limit its applicability to detect conflicting health information at large. First, rule base solutions are brittle and harder to generalize. Secondly, their solution requires a prohibitively expensive amount of annotation for each piece of advice and is hard to scale to other data. So, the next step in the advancement of HCD is to develop an end-to-end system which does not require granular levels of annotation and can work in a production environment on free-text inputs. The recent advances in transfer learning inspire us to investigate the scope of pre-trained language models to detect health conflicts automatically. Successful implementation of a transfer learning based solution can effectively address this critical challenge of automatically detecting inconsistent/conflicting health information at large. 

One of the main challenges with using pre-trained language models for HCD is that they require sufficient labeled data for fine-tuning and the HCD task is inherently resource-constrained. For example, given a set of $500$ health advice statements, there exists $(500)^2 = 250000$ candidate advice pairs from which a very small set will be conflicting --- making data collection extremely difficult. Given the limited number of health advice pairs presented in \cite{7917875, PREUM2017226}, this study explores methods of \textit{synthetic conflict augmentation} towards making HCD suitable for use in a transfer learning framework. After expanding the original HCD training set by 2x, we explore the task in the context of pre-trained language models. Finally, to verify prediction reliability, we  test if transformer-based predictions value the same information as human annotators when asked to identify the content most important to the prediction of a conflict. In summary, we considering the following research questions:

\paragraph{RQ1: How effective pre-trained language models are in detecting conflicting health information?} 
We provide thorough analysis of five different language models on the HCD task, investigating the challenges associated with understanding different conflict types. We additionally contrast our work with a related task, Natural Language Inference, which aims to classify pairwise inputs as entailed, contradictory, or neutral \cite{mnli_paper}. We investigate the relationship between notions of conflict and contradiction by applying a pre-trained NLI classifier to the HCD task. 

\paragraph{RQ2: Are synthetic health conflicts able to aid in the prediction of real health conflicts?}
% RQ3 measures the effectiveness of the human-in-the-loop synthetic data augmentation technique
While synthetic conflicts are shown to be linguistically and semantically valid, they are often factually incorrect and contain distributional drift in terms of style. So, we explore the capability of synthetic conflicts to aid in the prediction of real conflicts. \\

\paragraph{RQ3: Do language models agree with humans when  identifying features most influential to conflict detection?} 
The interpretability of AI systems is crucial to integration of machine learning in healthcare systems \cite{XAI}.  We use a popular model interpretability tool, Captum \cite{kokhlikyan2020captum}, to identify the input tokens our top performing language model attributes as being most important to the prediction of a given output. We wish to explore if language models pay attention to known conflict indicators such as the conflict topic and label-specific semantic phrases (LSSPs) (e.g. temporal or conditional clauses that result in a temporal or conditional conflict).

% Each advice pair displayed are comments from either verified physicians or registered nurses.

% \begin{table*}[htbp]
% \renewcommand{\arraystretch}{1.2}
% \begin{tabularx}{\linewidth}{ X X l l}

% \toprule
% \textbf{Medical Professional A}  & \textbf{Medical Professional B} & \textbf{Post Topic} & \textbf{Conflict Type}\\ 
% \midrule
% For clarity's sake, are you saying you took 12x500 mg over the course of 23 hours or 11 hours? If it's the former then that is fine. If it's the latter it may be worth getting checked as it's double the recommended dose. & Assuming the doses were spaced out evenly then concentrations would not have reached toxic levels at any point (see calculation below if you're interested#). & Fever Medication & Direct \\
% If the left breast has been getting larger and developing those veins recently, and if you feel firmness of the left breast that is different from the right, then I would get it checked out by your PCP or GYN. If they have always been different, then it’s just normal & They’re normal. Breasts (like eyebrows) are “sisters, not twins” - no one has perfectly symmetrical breasts. & Breast Cancer & Conditional \\
% \bottomrule
% \end{tabularx}
% \caption{Conflicting health information found on reddit.com/r/AskDocs --- A social media forum where anyone can get medical advice from a professional whose medical certifications are verified by the forum moderators. We show that suggestions from medical professionals can be conflicting. }  
% \label{social_media_data}
% \end{table*}

%% file: sections/1ProblemFormulation.tex
Inspired by the definition presented in \cite{7917875}, we define the input to a health conflict detection (HCD) task as a pair of health advice, $(W_1, W_2)$ each with a common topic or object of an advice, $t$. 
% For example, the conditional conflict sample ($W_1, W_2$) in Table \ref{dataset} has the common topic \textit{cruciferous vegetables}. 
HCD is a multiclass, multilabel classification problem. Each advice pair contains one or more of the following labels.

\noindent \textbf{Direct Conflict: } Occurs when the topic of a health advice pair has opposite polarity. If a piece of advice $W_i$ suggests the user take action $a$, a direct conflict occurs when another piece of advice $W_j$ suggests the user against taking action $\lnot a$. In Table \ref{dataset}, our direct conflict example has topic \textbf{snack}. We see how advising fruit as a healthy snack directly conflicts with the suggestion to limit sweet snacks. \\
\textbf{Temporal Conflict: } Occurs when a health advice pair disagrees about \textit{when} to take some action $a$. In Table \ref{dataset}, our temporal example agrees one should drink liquids, but disagrees as to when consumption should occur. \\
\textbf{Sub-Typical Conflict: } Occurs when the topic in reference to the suggested action disagree in type. In Table \ref{dataset}, we see that Advice 1 supports milk consumption while Advice 2 speaks negatively only in regards to \textit{full-fat} milks. \\
\textbf{Conditional Conflict: } Occurs when a conflict's occurrence depends on some condition. In Table \ref{dataset}, our conditional example displays how Advice 1 and Advice 2 are only conflicting if the user is taking Coumadin. \\
\textbf{Non-Conflict: } Occurs when two pieces of advice are non-conflicting. \\

We note that this list of conflict types is non-exhaustive, as we do not explore the quantitative and cumulative conflicts defined in \cite{PREUM2017226}, due to limited available annotation. Such conflicts may be explored in future works.

%% file: sections/3Dataset.tex
% assuming this was meant to be commented
% --real dataset one from preclude
% --real dataset that we collected and annotated following the procedure described in preclude
% --synthetic dataset

\begin{table}[!h]
\centering 
\begin{tabular}{ccccc}
\toprule
\textbf{Conflict Type} & \textbf{Augmentation Strategy} & \textbf{F1}  \\
\midrule
      & None             & 0.67 \\
      & Paraphrase       & 0.66  \\ %59
 Direct                  & Text Generation  & 0.62 \\
      & Back Translation & 0.65 \\
     & Aggregated Set    & 0.63 \\
\midrule
 & None                  & 0.15 \\
 & Paraphrase            & 0.08  \\
Sub-Typical              & Text Generation & 0.19\\
 & Back Translation.     & 0.09 \\
 & Aggregated Set        & 0.13 \\
\midrule
 & None                  & 0.40 \\
 & Paraphrase            & 0.25  \\ % 
Conditional              & Text Generation  & 0.25 \\
 & Back Translation      & 0.14 \\ %
 & Aggregated Set        & 0.38 \\
\midrule 
    & None               & 0.69 \\
    & Paraphrase         & 0.70  \\
Temporal                 & Text Generation  & 0.57\\
    & Back Translation   & 0.59 \\ % 0.64
    & Aggregated Set     & 0.69 \\
\bottomrule
\end{tabular}
\caption{The F1 score of DeBERTa-v3 when trained using additional data using various popular augmentation strategies. These results highlight the complexity involved in augmentation of pair-wise multi-sentence inputs and the need for more advanced augmentation algorithms. } 
\label{Augmentation_Experiment}
\end{table}

The following section is organized as follows. First, we provide an overview of existing HCD data. We then discuss our attempts at performing data augmentation on existing HCD samples. Next, we discuss how new data was collected. Finally, we detail our human-in-the-loop augmentation approach, which alleviates problems faced during our attempts at data generation and augmentation. Additional dataset details are included at the end of the section. 

% \subsubsection {Real Datasets}
% \textcolor{blue}{Two types/sets of real advice data: (i)PreCluDe, (ii) new online health data}

% Due to the nature of the HCD problem, textual health conflicts do not appear frequently in the wild, making data collection difficult. While the original HCD dataset introduced in \cite{7917875} uses all data from real health apps and health websites, this constraint makes it difficult to collect enough samples to produce a generalizable NLP system. To overcome this challenge we introduce  \textit{Synthetically Generated Conflicts} into the training set, doubling the number of samples in the original HCD training set. We describe in detail both the real and synthetic components of this dataset. 
% \label{section:eval_dataset}

\paragraph{Real Health Advice Dataset from \textit{PreCluDe}}
% \paragraph{(i) Collecting Real Health Advice Samples for HCD}
% Prior works on HCD collect data samples consisting of real health advice text collected from authentic health websites, health apps and drug usage guidelines for prescription medications used to treat chronic diseases \cite{preum2017preclude2, preum2018corpus}. 
% The dataset was constructed from two prior public datasets on textual health advice \cite{preum2018corpus, preum2017preclude2}. 
% It contains three categories of data. First, it contains 1156 pieces of advice statements on five general health topics (e.g., weight loss, pregnancy, nutrition, exercise, digestive health) and two common chronic diseases including diabetes and anemia. This set of advice was collected from eight mobile health apps (including four Android and 4 iOS apps), and four authentic health websites. Second, it contains 1124 pieces of advice statements on 34 chronic diseases from disease specific health websites. These diseases include common chronic conditions in the USA, including but not limited to, diabetes, heart disease, depression, hypertension, chronic kidney disease, obesity, arthritis, backpain, alcohol and substance abuse disorder. Finally, our real health advice dataset also has 1005 advice statements from drug usage guideline documents corresponding to the 90 medications that are commonly used to treat the 34 chronic diseases mentioned above. The list of diseases and medications are selected based on a set of real prescriptions for multi-morbidity. 
The PreCluDe dataset contains 3285 pieces of advice statements from authentic health websites, health apps, and drug usage guidelines. The data covers general health topics, 34 chronic diseases, and their relevant prescription medications collected from authentic and reliable health websites and popular health apps. The list of diseases and medications are selected based on a set of real prescriptions for multi-morbidity. These advice statements can be single sentence or multi-sentence. Each of these advice statements were annotated for topics of advice (e.g., advice on a food item, drug-drug interaction, drug-food interaction, exercise). In addition each advice was annotated based on the polarity of the advice with respect to each topic. Then potential pairs of conflicting advice statements were generated based on common topics. Each of these potential pairs of conflicting advice statements were then annotated by at least three reviewers to determine whether they are conflicting and the type of potential conflicts (if any). The details of this annotation process are described in multiple published literature \cite{preum2017preclude2, preum2018corpus}. The PreCluDe dataset is the source of the 2006 \textit{real} conflicting advice pairs utilized in this study. 

\paragraph{Synthetic Dataset Generation using Human-in-the-loop Data Augmentation} 
In attempt to expand the size of existing HCD data, we first explored various automated methods of textual augmentation. The result of these experiments can be found in Table \ref{Augmentation_Experiment}. Our first experiment employed PEGASUS \cite{pegasus} for paraphrase generation. Given some Advice 1, we paraphrase it's corresponding Advice 2 to generate a new sample. Unfortunately, this method was prone to the removal or alteration of the conflict topic, producing incorrect samples. We additionally tried a similar approach using Back Translation between German and English, but little textual diversity was observed. We also explored a text generation approach using GPT-Neo \cite{gpt-neo}, where we employed in-context learning to generate new samples, with prompt structure inspired by GPT3-MIX \cite{gpt3-mix}. Specifically, given a prompt and a small set of example advice pairs, GPT was asked to then generate the Advice 2 for a provided Advice 1. Unfortunately, this method failed to produce consistent coherent outputs. In each experiment we double the size of the dataset (thus generating n = 1398 samples) using the described augmentation methods. Note that we additionally report a result denoted Aggregated Set, which is an evenly distributed mix of all three augmentation approaches.

\paragraph{New Online Health Advice Dataset}
To expand upon the existing HCD dataset with additional real-world samples, we scraped 4 popular health information websites including WebMD\footnote{https://www.webmd.com/a-to-z-guides/common-topics}, MedLine\footnote{https://medlineplus.gov/healthtopics.html}, Covid protocols\footnote{https://covidprotocols.org/en/chapters/home-and-outpatient-management/}, and CDC\footnote{https://www.cdc.gov/DiseasesConditions/}. We collected a total of 47,600 candidate advice statements from a wide variety of health topics. Three annotators were then tasked with annotating a random sample of 5393 advice statements, each answering the following questions: 1) Does this statement contain health advice? 2) What are the topics of the advice? 3) What is the polarity of each topic in a piece of health advice? This allowed us to generate potential conflicting advice pairs by considering samples with common topics containing opposite polarities.

% See our data flow diagram in Figure \ref{data_flow_v3} for additional data collection details.
The annotators found 2715 samples which were in fact health advice, and thus eligible for inclusion in a conflicting advice pair. Unfortunately, very few useful organic conflicts were found when pairing together samples with opposite topic polarities. This occurred because 1) there is a wide variety of topic entities annotated, leaving few opportunities for sample pairing, 2) many topics have inherent static polarity (e.g. cancer is always negative), which makes finding a conflict for such samples difficult. We thus concluded that automatic conflict generation using randomly-sampled, real-world text was not a feasible strategy to \textit{significantly} increase the dataset size.

\paragraph{Human-in-the-loop Data Augmentation} 

The failure of the two aforementioned approaches to data generation inspired us to take a human-in-the-loop approach for data augmentation using mTurk. This approach allows us to generate training samples which encode the relevant linguistic phenomena associated with each conflict type, even if the style and prose of human-written samples does not perfectly reflect real-world data. Although the synthetic data generated in this process is most often not clinically accurate, they are helpful to capture the linguistic phenomena of conflicting textual information. 
In this study, we recruit 88 master-qualified mTurk workers. Additionally, we only recruited US-based annotators with a minimum 90\% acceptance rating. To generate a synthetic advice conflict, an mTurk worker (Turker) is provided with a real health advice statement, the topic of the advice, and instructions on how to write samples for each conflict type. For example, a Turker may be prompted with the advice statement "Drink less alcohol." If asked to write a conditional conflict, a valid synthetic sample would be "Drink less alcohol \textit{if you suffer from heart disease}". Full details of the data collection process are depicted in Figure \ref{data_flow_v3}.

Of the 2715 newly collected health advice statements deemed useful, an additional round of human review extracted 649 advice statements most suitable for augmentation. Each sample was annotated for each conflict type, and then \textbf{ verified by at least one research assistant at the authors institution.}  The resulting dataset after sample verification contains 1897 unique advice statements from Amazon's mTurk platform \footnote{Please refer to the supplementary materials for further details on mTurk annotation instructions.}. We display the distribution of health domains included in the synthetic dataset in Table \ref{topic_distribution}.

% \begin{figure}
%     \centering
%     \includegraphics[scale=0.42]{Figures/topicDistribution_v2 (1).pdf}
%     \caption{Distribution of health advice domains in the newly collected data. }
%     \label{topic_distribution}
% \end{figure}

\begin{table}
\centering 
\begin{tabular}{lc}
\toprule
\textbf{Advice Topic} & \textbf{Number of Samples}  \\
\midrule
Adhd  &  19\\
Allergies  &  52\\
Alzheimers  &  59\\
Arthritis  &  115\\
Asthma  &  73\\
% Covid-19  &  1\\
Diabetis  &  51\\
Druginfo  &  44\\
Healthtopics  &  32\\
Hypertension  &  86\\
Pregnancy  &  92\\
Other & 26\\
\bottomrule
\end{tabular}
\caption{Distribution of health advice domains in the newly collected data. } 
\label{topic_distribution}
\end{table}

\subsubsection{Dataset Details}
In total, there are 2006 real annotated advice pairs, and 1897 synthetic pairs. Our training set contains 2825 samples, comprised of a mix of real and synthetic advice pairs. We evaluate our model on two test sets: One with entirely real-world data, and another which contains only synthetic advice pairs. The full data distribution is shown in Table \ref{data_disribution}. The ratio of single-sentence advice to multi-sentence advice is 3.07. The length of a multi-sentence advice can vary from (2, 11) sentences. On average, one health advice is 1.4 sentences long, containing close to 25 words. Given that HCD is a pairwise task, each input thus requires modeling of about 2.8 sentences, demanding long-range dependency modeling.

%% file: sections/4Methods.tex
\subsection{RQ1: How effective pre-trained language models are in detecting conflicting health information?}
To investigate the strengths and weaknesses of pre-trained language models on the HCD task, we aim to first quantify performance when fine-tuning transformer-based models for HCD. 
% To illustrate, consider the special tokens used by BERT, where the input to the network is $[CLS] \: Advice \: 1 \:  [SEP] \: Advice \: 2 \:  [SEP]$. A model then computes a contextual embedding for each token, and the special $[CLS]$ token is then fed to a linear classification head for conflict prediction. This is the general paradigm used in each experiment. 
We fine-tune 5 different models for each architecture in our experiments. Four out of these five models are 1-vs-all classifiers for each conflict type, with the fifth being a multi-label classifier.  We then report results on two different test sets, each containing exclusively real or synthetic data. These experiments will identify (i) which conflict-types are most and least challenging to detect and (ii) allow us to quantify the difficulty level in predicting real vs synthetic test samples. 

We additionally explore the relationship between conflict detection and the relevant NLP task of Natural Language Inference, which includes \textit{contradiction detection}. To do so, we apply a DeBERTa-v3 model fine-tuned on the MNLI dataset \cite{mnli_paper} and consider contradictory predictions to be conflicts in our 1-vs-all HCD setting. This experiment aims to highlight the differences between conflicts and contradictions and motivate the need for HCD-specific algorithms. We additionally experiment with intermediate fine-tuning \cite{IFT} on MNLI, before fine-tuning on HCD. This experiment aims to explore if leveraging the subtle relationships between MNLI and HCD provides any boost in predictive performance. 

% We additionally investigate data-specific challenges faced by the BERT model. 1) Are longer advice statements more difficult to understand? To answer, we analyze performance when accounting for both token and sentence length of advice statements. 2) Are health conflicts about certain topics more difficult to understand? In this experiment, we identify topics which are difficult to predict. 
% 3) Can we predict out-of-distribution health topics? In this experiment, we construct a train/test split where the health advice topics in each set are disjoint. 

\subsection{RQ2: Are synthetic health conflicts able to aid in the
prediction of real health conflicts?} 

We have thus far motivated the need for better health conflict detection solutions, which will require more data to train reliable end-to-end models. Thus, RQ2 measures the effectiveness of the human-in-the-loop synthetic data augmentation technique.
% we have succeeded in finding a reliable solution to health conflict data augmentation. 
This experiment evaluates the predictive power of synthetic training samples on real-world health conflicts. To evaluate, we train 3 separate classifiers on each conflict type, using real-only, synthetic-only, and real+synthetic training data respectively.

\subsection{RQ3: Do language models agree with humans when  identifying features most influential to conflict detection?} 

\begin{figure}
    \centering
    \includegraphics[width=\textwidth/2,height=\textheight,keepaspectratio]{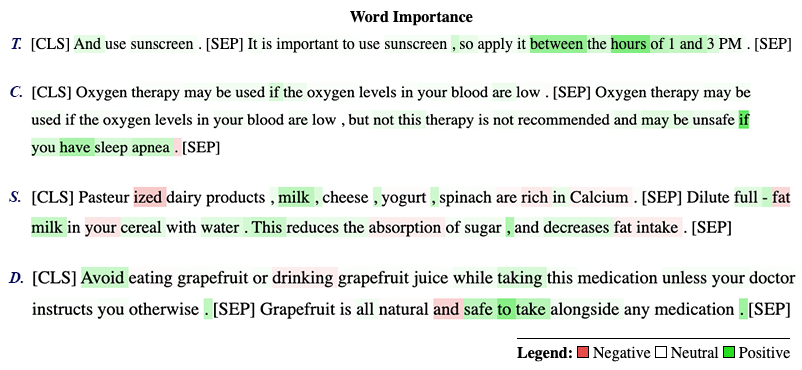}
    \caption{Visualizing sample outputs of token importance attributions from the Captum library. Each row is labeled with it's conflict label: \textbf{T}emporal, \textbf{C}onditional, \textbf{S}ub-typical, and  \textbf{D}irect.  }
    \label{fig:captum}
\end{figure}

% As described in the \textbf{Problem Formulation} section, each conflict is annotated \textit{with respect to a topic.} Thus, it is of interest to note if transformer models pay attention to the conflict topic or relevant semantics of the advice to make the predictions. 

With the long-term goal of deploying a scaleable HCD system to alert health information consumers of potential conflicts, we aim to explore the reliability of transformer-based model's by taking a low-level look at their predictions on the HCD task. To do so, we use the Layer Integrated Gradients algorithm \cite{IntegratedGradients} provided by Captum \cite{kokhlikyan2020captum}, to quantify the attribution of each input token to the prediction of the conflict label. 
% \textcolor{blue}{should discuss more about this tool and how it finds the part where BERT is looking at. what is numeric attribution score?}
Specifically, Captum allows us to produce a numeric attribution score for each input token. This enables visualization of token importance by toggling the opacity of highlighted tokens based on the magnitude of their attributions. An example Captum output for each conflict type is shown in Figure \ref{fig:captum}. Tokens highlighted in Green contribute positively to the prediction of the positive class. Tokens highlighted in red contribute positively towards the prediction of the negative class. In this experiment, we employ 3 human annotators experienced in health conflict detection to answer the following two questions about the tokens Captum identifies as most important to the prediction of the positive class. 

\begin{enumerate}
    \item Does the model identify the topic as important to classification? 
    \item Does the model identify the correct Label-Specific Semantic Phrase (LSSP) as important to classification? Where LSSPs are defined as the topic-altering tokens which generate a specific conflict. 
\end{enumerate}

In other words, the annotator is asked to look at a given sample, gather their understanding of the correct Topic and LSSP, and confirm their intuitions align with the transformer model by observing which tokens DeBERTa-v3 highlights in green. Given the importance of model interpretability in making health-related decisions \cite{vellido2020importance, dai2020ginger}, this experiment will verify if transformers are making judgements which align with human intuitions.
%  about different types of conflicts.
In this experiment, each annotator looks at 10 samples from each class, from each test set (80 samples in total). All samples were selected randomly from the set of correct predictions made by the DeBERTa-v3 model. Each annotator provides a binary response (Yes/No) for each question, and the average score is reported.

\begin{table}[!h]
\centering 
\begin{tabular}{lccc}
\toprule
\textbf{Conflict Type} & \textbf{$\textbf{NLI}_{OTS}$} & \textbf{$\textbf{NLI}_{FT}$} & DeBERTa-V3 \\
\midrule
Direct           & 0.53 & 0.69 & 0.75\\ 
Sub-Typical      & 0.12 & 0.29 & 0.36\\
Conditional      & 0.08 & 0.69 & 0.53\\
Temporal         & 0.05 & 0.81 & 0.70\\ 
\end{tabular}
\caption{Experiments exploring the relationship of HCD and NLI. \textbf{$\textbf{NLI}_{OTS}$} identifies how well an off-the-shelf (OTS) NLI model predicts conflicts.  \textbf{$\textbf{NLI}_{FT}$} highlights the impact of intermediate fine-tuning (FT) DeBERTa-V3 on the MNLI task. i.e. this model is first fine-tuned to MNLI, then fine-tuned to HCD. DeBERTa-V3 is the reference HCD score with no NLI training. All results are mean F1 score over 3 experimental trials. } 
\label{NLI}
\end{table}   

\section{Evaluation Setting}
% \subsubsection{Performance Metrics}
To ensure results are robust to random weight initializations, we run each experiment 3 times, each with a different random seed. Each reported result is the macro average of all experiments. For each individual experiment, we compute the F1 score of the positive class in a 1-vs-All setting or the weighted F1 in the multi-label setting. All metrics are computed using the scikit-learn package \cite{pedregosa2011scikit}. We note that we are unable to perform cross-validation due to the fact that a given advice statement may be paired with multiple other advice statements. Thus, performing cross validation runs the risk of input memorization as has been observed in similar pairwise inference tasks \cite{Gururangan2018AnnotationAI} \cite{herlihy2021mednli}. This issue additionally inhibits our ability to construct a separate validation set. For example, we cannot simply take 200 samples out of the real test set and create a validation set as many pieces of advice are used in multiple conflicting advice pairs --- which would cause data leakage. Thus, hyperparameter optimization was considered out-of-scope of this work but should be explored in future works. 

Experiments using static embedding methods, namely Sentence-BERT (SBERT) \cite{reimers-2019-sentence-bert} and GloVE \cite{pennington2014glove} each function by first computing a single embedding vector for each piece of advice. Next, the embeddings for Advice 1 and Advice 2 are concatenated and fed into a linear classification head using scikit-learn. In these experiments, only the classification head undergoes training. Each dynamic contextual embedding model (BERT, Bio+Clinical BERT, RoBERTa, DeBERTa-v3) is fine-tuned useing the pair-wise clasification paradigm outlined in \cite{bert} for a fixed 5-epochs with the AdamW \cite{AdamW} optimizer set with learning rate = $2e-5$ and a weight decay = 0.01 to combat overfitting. We use the standard Cross Entropy loss for each task. In the case of 1-vs-All experiments, the loss is weighted using the distribution of the training labels to account for class imbalance.

% \subsubsection{Experimental Hyperparameters}
% All transformer models were trained using the Huggingface Transformers library \cite{wolf-etal-2020-transformers}. All models were fine-tuned for 5 epochs using AdamW \cite{adamw} optimizer with a learning rate = $2e-5$, batch size = 16, and weight decay = 0.01. To account for class imbalance in the 1vsAll setting, we use a weighted cross entropy loss where weights are balanced using the training set label distribution. Our random forest baseline similarly used a balanced class weight objective. We note that no hyperparameter optimization was done for any model as this is outside the scope of this work. 

% \subsubsection{Baseline Models}

% \noindent \textbf{RF}: Random forest classifier using TF-IDF feature of the concatenated advice pair.  

% \noindent\textbf{BERT}: This baseline uses the default pre-trained `bert-base-uncased' model provided by Huggingface \cite{wolf-etal-2020-transformers}. Classification is performed by first getting a contextual embedding given our pair-wise advice input. Next, we use the [CLS] token as described in \cite{bert} as input to a linear classifier, which is shown in Figure \ref{model_diagram}.

% \noindent \textbf{Bio+Clinical BERT}: This baseline aims to test if HCD benefits from medical domain-specific knowledge by using contextual embeddings from Bio+Clinical BERT \cite{alsentzer-etal-2019-publicly} --- a BERT-style architecture trained on scientific/clinical texts. 

% \subfile{sections/Evaluation_Setting}

%% file: sections/5Results.tex
\newcommand{\ra}[1]{\renewcommand{\arraystretch}{#1}}
\begin{table*}\centering
\ra{1.2}
\begin{tabular}{@{}cccccccccccccccr@{}}\toprule
& \multicolumn{2}{c}{Direct} & \phantom{abc}& \multicolumn{2}{c}{Sub-Typical} &
\phantom{abc} & \multicolumn{2}{c}{Conditional} & \phantom{abc} & \multicolumn{2}{c}{Temporal} & \phantom{abc} & \multicolumn{2}{c}{Multilabel}\\
\cmidrule{2-3} \cmidrule{5-6} \cmidrule{8-9} \cmidrule{11-12} \cmidrule{14-15} 
& $F_r$ & $F_s$  && $F_r$ & $F_s$ && $F_r$ & $F_s$  && $F_r$ & $F_s$ && $F_r$ & $F_s$ & \textbf{\textit{Avg}}\\

\midrule           
Random Guess        & 0.39 & 0.28  && 0.24 &0.25  && 0.40 & 0.33  && 0.24 &0.30 && 0.35 &0.32 &0.31 \\
TFIDF+RF            & 0.07 & 0.36  && 0.00 &0.00  && 0.00 & 0.25  && 0.21 &0.36 && 0.03 &0.18 & 0.14\\
% \arrayrulecolor{black!30}\midrule

GloVE               & 0.51 & 0.34  && 0.21 &0.19  && 0.58 & 0.46  && 0.40 &0.64 && 0.42 &0.43 & 0.41\\
SBERT               & 0.34 & 0.55  && 0.17 &0.25  && 0.31 & 0.50  && 0.49 &0.72 && 0.37 &0.51 & 0.42\\
\midrule 
BERT                & 0.38 & 0.73  && 0.28 &0.18  && 0.44 & \textbf{0.86}  && 0.53 & \textbf{0.91} && 0.40 &0.70 & 0.54 \\
B+C BERT     & 0.36 & 0.70  && 0.24 &0.23  && 0.45 & 0.84  && 0.66 &0.89 && 0.34 &0.70 & 0.54\\
RoBERTa             & 0.51 & 0.77  && 0.30 &0.39  && \textbf{0.58} & 0.85  && \textbf{0.72} &0.90 && \textbf{0.56} & \textbf{0.79} & 0.63\\ 
DeBERTa-v3          & \textbf{0.75} & \textbf{0.85}  && \textbf{0.36} & \textbf{0.71}  && 0.53 & 0.83  && 0.70 & \textbf{0.91} && 0.51 &0.71 & \textbf{0.68}\\ 
\bottomrule 
\end{tabular}
\caption{Average F1 score across 3 experimental trials using all training data to predict each conflict type on the real ($F_r$) and synthetic ($F_s$) test sets respectively.} 
\label{Results_Table}
\end{table*}

\subsection{RQ1: \textbf{How effective pre-trained language models are in detecting conflicting health information?}}

\subsubsection{Model Analysis}
We report the performance of 5 different transformer-based classifiers in Table \ref{Results_Table} as well as a random guess, lexical feature, and word embedding-based baseline. The lexical model, which uses TFIDF \cite{tfidf} for feature generation with a class-weighted Random Forest classifier \cite{RF}, under-performs random guess on 90\% of experiments across all conflict types. This is expected as TFIDF generates features based on word count and frequency, ignoring the fact that the input text is comprised of two disjoint advice statements which need to be \textit{compared}. Additionally, we note that the Random Forest model was unable to overcome the severe class imbalance with TFIDF features, often  over-predicting the non-conflicting class for all labels.

The static embedding methods, SBERT and GloVE, provide an improvement over our lexical baseline while out-performing random guess in most experiments. However, static embedding methods are not constructed to perform pair-wise comparison of longer texts. Thus, all textual comparison must happen in the classification head, which is shown to be inferior to the dynamic embedding approaches shown in Table \ref{Results_Table}.

\begin{figure}
    \centering
    \includegraphics[scale = 0.21]{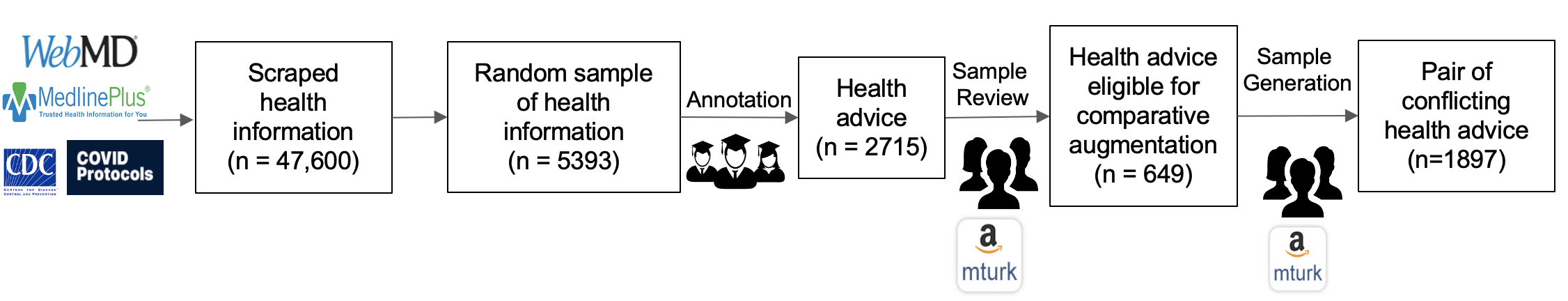}
    \caption{Data flow diagram depicting the data collection process.}
    \label{data_flow_v3}
\end{figure}

Transformer-based models such as BERT provide significant improvement over both TFIDF and static embeddings, as their contextual embedding approach permits direct comparison of two texts. The BERT baseline, however, is shown to be inferior to more modern transformer models such as RoBERTa and DeBERTa-v3, both of which outperform BERT on 80\% of experiments. From our experiments using Bio+Clinical BERT (B+C BERT), we show that BERT's performance issues are not due to BERT's lack of domain specific training data, as B+C BERT under-performs or matches BERT's performance on 70\% of experiments. This again follows intuition as, while this task is called \underline{health} conflict detection, the addition of knowledge from sources like biomedical corpora and clinical notes are intuitively not relevant to this task. Health advice contains common language alongside occasional references to medications or physiological effects (as shown in Table \ref{dataset}).  

Across all experiments for which scores were higher than random guess, results on the synthetic test set  outperform results on real health conflicts --- often by a significant margin. This confirms our understanding of synthetic data outlined in the dataset description - where we hypothesize that non-organic conflict data may be easier to understand as they often contain simpler and less diverse sentence semantics, using phrases specific to label definitions (as discussed in the \textit{Dataset} Section). 

\smallskip

\noindent \textbf{Label Specific Performance Analysis}

\begin{figure}
    \centering
    \includegraphics[width=8.5cm]{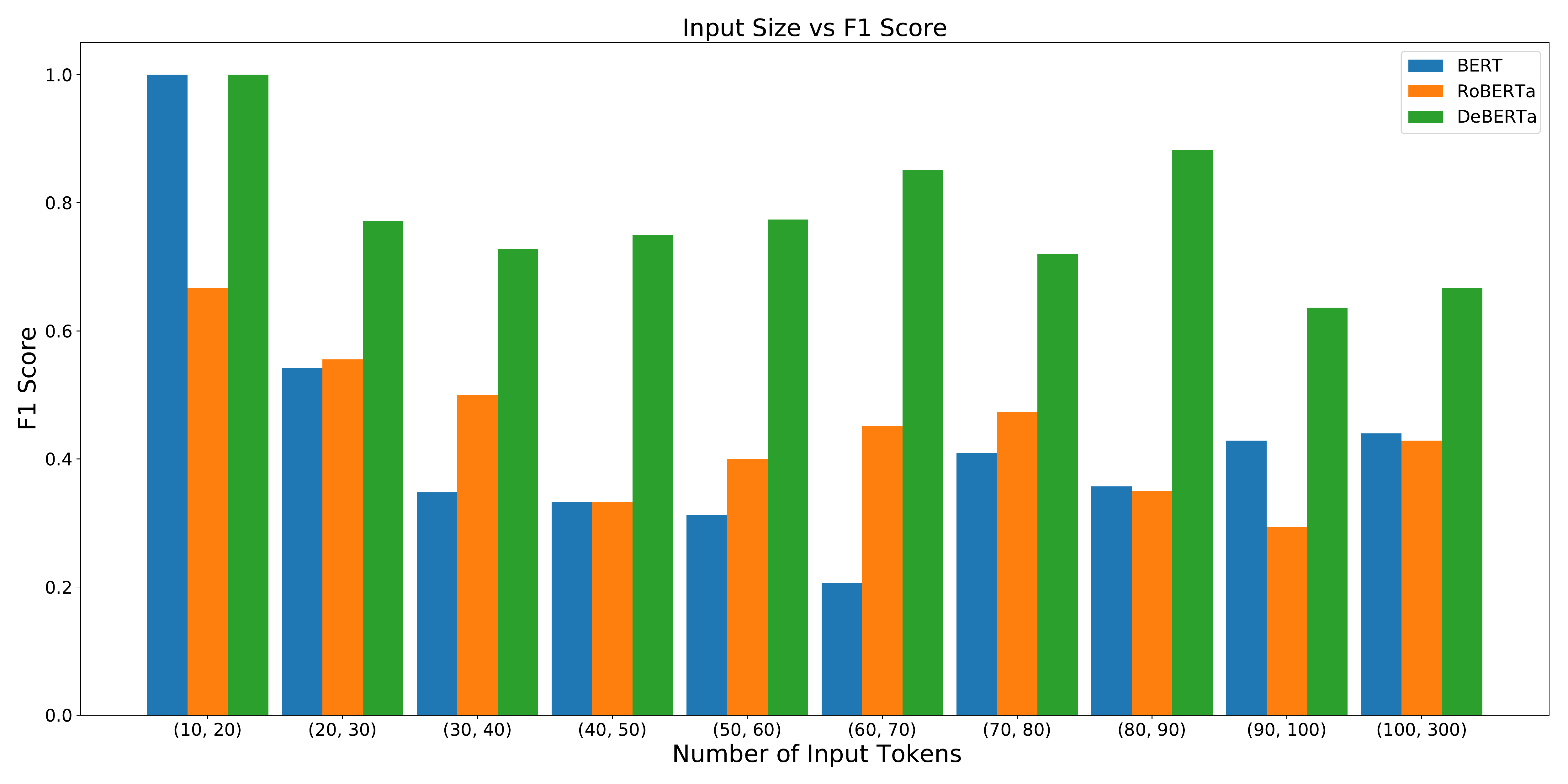}
    \caption{Plot visualizing the relationship between token length and model performance. The x-axis represents token number bins and the y-axis shows the corresponding F1 score on that data subset. Results were generated using all training data in a one-vs-all setting on Direct Conflicts. }
    \label{fig:length_vizualization}
\end{figure}

\smallskip
\noindent \textbf{Direct Conflicts:} Our results show there is a clear and impactful variance in the data distributions between real and synthetic direct conflicts, as performance on their respective test sets has the highest discrepancy across all conflict types. We additionally note the \textit{significant} improvement provided by DebERTa-v3 on direct conflict understanding. We believe the DeBERTa-v3 architecture is well-suited for this conflict type for the following reasons. Direct conflicts, by definition, will not contain overt textual clues in a single advice statement which facilitate better label prediction. This in contrast to temporal conflicts, for example, where a statement such as ``before bed" found in Advice 2 is a strong indicator that there \textit{may} be a temporal conflict independent of  Advice 1. Thus, in order to understand direct conflicts, long-range dependency modeling is required, where a common topic must be identified in both advice statements as well as their relative polarities. DeBERTa-v3 improves upon BERT and RoBERTa in a variety of ways that facilitate long-range dependency modeling. Specifically, DeBERTa-v3 uses (i) disentangled position embeddings, (ii) global position information, and (iii) replaced token detection pre-training, which has been shown to be a more efficient pre-training objective than those used by BERT and RoBERTa \cite{he2021debertav3}. We find evidence of DeBERTa-v3's improved ability to reason about pair-wise inputs in their paper where DeBERTa-v3 outperforms BERT on the pairwise Recognizing Textual Entailment task by an astounding 20.3 accuracy points \cite{he2021debertav3}. We visualize the relationship between input length and F1 score in Figure \ref{fig:length_vizualization}, where we observe how BERT and RoBERTa are much more sensitive to token length than DeBERTa-v3 is for direct conflicts.  

A common error made when predicting Direct Conflicts was the over-prediction of other conflict samples which contained opposite topic polarities alongside conditional or temporal constraints. For example, if Advice 1 has positive sentiment towards coffee yet Advice 2 expresses negative sentiment towards coffee \textit{only for pregnant women}, false positives tend to occur as the model is unable to understand that the opposing sentiments are tied to a conditional constraint. 

\textbf{Sub-typical Conflicts:} Our experimental results find that sub-typical conflicts are the most challenging to predict.  Similar to direct conflicts, sub-typical conflicts by definition have less-overt textual clues than conditional and temporal conflicts. However, unlike direct conflicts, which rely on understandings of polarity with respect to each topic, sub-typical conflicts often rely on world-knowledge for proper understanding. For example, the advice pair (``Don't eat cheese", ``Only some cheeses are bad for you, like Goat cheese") contains a sub-typical conflict regarding cheese type. It may be easy for a pre-trained language model to understand the topic (cheese) and sentence polarities (negative, negative), but challenging to know that "goat" is a type of cheese, and that the input isn't referring to the animal itself, but rather altering the topic-type. 

\textbf{Temporal and Conditional Conflicts:} Pre-trained language models find the most success in the prediction of conditional and temporal conflicts. This is particularly true of the synthetic test sets, where mTurk workers could add \textit{if} or similar conjunctions or temporal clauses to the end of the original advice statements to generate conditional or temporal conflicts, respectively. Thus such samples are less challenging for the model to classify correctly. As previously stated, these conflict types contain more consistent textual patterns that make their detection easier. Interestingly, these two conflict types are the only experiments where DeBERTa-v3 does not outperform other pre-trained language models. This is likely due to the lesser dependency on long-range dependency modeling required by these two conflict types. 

With these two conflict types, the presence of label-specific semantic phrases in non-conflicting samples can cause false positives. Additionally, our conditional classifier was stumped by samples such as "Don't exercise on an empty stomach" where the condition is expressed in an implicit, more compact form. That is to say, the model may have predicted it correctly had it been phrased "Don't exercise if you have an empty stomach". Problems such as these can be mitigated by collecting more diverse samples in future works. 

\textbf{Relationship to Natural Language Inference:} Table \ref{NLI} highlights the results of our experiments exploring the relationship between NLI and HCD. The $\mathbf{NLI}_{OTS}$ experiment tests how well an off-the-shelf (OTS) NLI model detects health conflicts. Our results show extremely poor performance from $\mathbf{NLI}_{OTS}$ on most conflict types, with moderate performance on Direct Conflicts. This result was expected as direct conflicts are semantically the most similar to NLI contradictions. However, nuanced conflict types like conditional, temporal and sub-typical have no relationship to the NLI output space, making this a near impossible task for an $\mathbf{NLI}_{OTS}$ model. For example, consider the instructions given to MNLI annotators when writing contradictions: Given a sentence 1, write a sentence 2 which is ``definitely incorrect about the event or situation" in sentence 1.   In the case where Advice 1 is \textit{Don't consume alcohol} and Advice 2 is \textit{Don't drink alcohol if you're pregnant}, we have a conditional conflict which is not at all related to the requirements for an MNLI contradiction. 

The $\mathbf{NLI}_{FT}$ experiment highlights how NLI may be useful to some conflict types as an intermediate fine-tuning task. Specifically, Temporal and Conditional conflicts benefited significantly from this pre-training strategy. However, given the previous discussion on the difference in definitions between conflict and contradiction, we suppose the performance increase is due to a combination of the following reasons: 1) The introduction of DeBERTa-v3 to a pairwise inference task and 2) These conflict types have overt textual clues, making them easier to classify. In other words, we can classify more of the simpler sample types with additional pair-wise inference pre-training. However, this had no effect on the comparison of complex sample types such as Direct and Sub-Typical. 

% \begin{table}[!h]
% \centering 
% \begin{tabular}{lc}
% \toprule
% \textbf{Conflict Type} & \textbf{F1}\\
% \midrule
% Direct      & TBA \\
% Sub-Typical & TBA \\
% Conditional & TBA \\
% Temporal    & TBA \\ 
% Multi-Label & TBA\\ 
% \bottomrule
% \end{tabular}
% \caption{OOD Prediction} 
% \label{RQ3}
% \end{table} 

\subsection{RQ2: Are synthetic health conflicts able to aid in the prediction of real health conflicts?}
A core goal of this paper is to expand the size of the original HCD datasets presented in a scaleable manner as collecting real data on conflicting advice is prohibitively expensive \cite{7917875, PREUM2017226}. We run experiments using different combinations of training data to quantify the contribution of synthetic samples on the real-world test set. Table \ref{RQ3} displays the results of this experiment. 
The synthetic data helped all transformer based models to classify conditional and temporal conflicts. 
% We find that all models consider synthetic data useful in the prediction of conditional and temporal conflicts. 
This validates our hypothesis that, even when there exists severe stylistic drift \footnote{Additional discussion of stylistic drift can be found in the limitations section. } between real and synthetic data, models can benefit from examples of LSSP's, regardless of their context. All models additionally benefit from synthetic sub-typical conflicts. In fact, \textbf{each model found synthetic data more useful than real data when predicting sub-typical conflicts.}   

However, the synthetic data was distracting when predicting direct conflicts for BERT and RoBERTa, while DeBERTa-v3 found synthetic data useful. This can be explained by DeBERTa-v3's superior long-range dependency modeling (as discussed in RQ1). Consider that the mean token lengths for a synthetic direct conflict and real direct conflict are 55 tokens and 86 tokens, respectively. In terms of sentence length, this on average translates to real direct conflicts being  1 full sentence longer than synthetic conflicts. Thus, synthetic direct conflicts are significantly shorter and perhaps not helpful in aiding BERT and RoBERTa to resolve the long-range dependency in real world data.  We find that DeBERTa-v3 is, in general, able to best utilize synthetic samples as they improve DeBERTa-v3's performance on real test data across all conflict types.

\begin{table}
\centering 
\begin{tabular}{llc}
\toprule
\textbf{Conflict Type} & \textbf{Topic} & \textbf{LSSP} \\
\midrule
Direct      & 0.18 & 0.75 \\
Sub-Typical & 0.66 & 0.63 \\
Conditional & 0.45 & 0.95 \\
Temporal    & 0.2 & 0.86 \\ 
\bottomrule
\end{tabular}
\caption{Experimental results from RQ3. Scores in the column \textbf{Topic} confirm if DeBERTa identifies the conflict topic as important. Scores in the column \textbf{LSSP} confirm if DeBERTa is recognizing tokens which are relevant to the label type. } 
\label{RQ2}
\end{table}

\subsection{RQ3: Do language models agree with humans when  identifying features most influential to conflict detection?}

Table \ref{RQ2} displays the results of our experiment in which three human annotators are asked to confirm if DeBERTa-v3 identifies what tokens are most important to understand a conflict. For all experiments, the \textbf{Topic} score displayed in Table \ref{RQ2} is defined the same --- the human annotator is asked to confirm if the model has attributed a high score to the conflict topic. LSSP's however are defined per conflict type. In the following sub-sections, we define the LSSP for each conflict type and analyze the results. 

\subsubsection{Direct Conflicts: } In this experiment, LSSP points were awarded to samples where DeBERTA-v3 identified relevant polarity tokens associated with each conflict  topic. In Figure \ref{fig:captum}, we see that the topic \textit{Grapefruit} is not emphasized --- which is surprising given the importance of conflict understanding in humans. However, the model correctly attributes "Avoid" and "safe to take", the two opposing polarity phrases, as being more important to classification. On average, Topics were attributed highly in 18\% of samples, while LSSP's were attributed highly in 75\% of examples.

\subsubsection{Subtypical Conflicts: } In this experiment, the model got points for LSSP identification if tokens which altered the type of the conflict topic were given high attribution scores. Our results found that 66\% of samples identified Topic and 63\% of samples identified LSSP's.
% , with only 20\% identifying both in the same sample. 
In the sub-typical example provided in Figure \ref{fig:captum},  the topic "Milk" is identified in both Advice 1 and 2, while the topic altering tokens "full -" are highlighted in Advice 2. Given the need to identify topic to understand sub-typical conflicts, it is expected that the topic attribution rate is  high for this conflict type. 

\subsubsection{Conditional and Temporal Conflicts} In these two experiments, LSSP attribution was credited when the model highlighted either conditional or temporal clauses respectively. For example, in Figure \ref{fig:captum},  the conditional example gives strong attribution to the statement "if you have sleep apnea". Additionally, in the temporal example, we observe strong attribution given to "between the hours of 1 and 3 pm". LSSP's for these two label types are shown to be easy to understand as conditional and temporal identify LSSP's 95\% and 86\% of the time respectively. However, topic attribution for these two conflict types are low, which is expected given the strong textual patterns associated with LSSP's for these conflict types.

\begin{table}
\centering 
\begin{tabular}{llccc}
\toprule
\textbf{Conflict Type} & \textbf{Training Set} & \textbf{$B_{F1}$} & \textbf{$R_{F1}$} & \textbf{$D_{F1}$} \\
\midrule
Direct & Real             & \textbf{0.49} & \textbf{0.62} & 0.67 \\
Direct & Synthetic.       & 0.07 & 0.20   & 0.24 \\
Direct & Real + Synthetic & 0.38 & 0.51   & \textbf{0.75} \\
\midrule
Sub-Typical & Real               & 0.18 & 0.13 & 0.15 \\
Sub-Typical & Synthetic          & \textbf{0.31} & 0.35 & 0.33 \\
Sub-Typical & Real + Synthetic   & 0.28 & \textbf{0.39} & \textbf{0.36} \\
\midrule
Conditional & Real             & 0.24 & 0.42 & 0.40 \\
Conditional & Synthetic        & 0.19 & 0.27 & 0.26 \\
Conditional & Real + Synthetic & \textbf{0.44} & \textbf{0.58} & \textbf{0.53}\\
\midrule 
Temporal & Real.            & 0.51 & 0.65 & 0.69 \\
Temporal & Synthetic        & 0.44 & 0.52 & 0.57 \\
Temporal & Real + Synthetic & \textbf{0.53} & \textbf{0.72} & \textbf{0.70} \\
\bottomrule
\end{tabular}
\caption{The F1 score of BERT, RoBERTa, and DeBERTa ($B_{F1}$, $R_{F1}$, $D_{F1}$ respectively) on the HCD real test set using real-only, synthetic-only, and all training data sets.} 
\label{RQ3}
\end{table}

% \begin{table}[!h]
% \centering 
% \begin{tabular}{lc}
% \toprule
% \textbf{Conflict Type} & \textbf{F1}  \\
% \midrule
% Direct           & 0.72 \\ 
% Sub-Typical      & 0.18 \\
% Conditional      & 0.45 \\
% Temporal         & 0.52 \\
% \end{tabular}
% \caption{Performance of BERT on HCD with intermediate fine-tuning on MNLI. I.e. the model was first trained on MNLI, then the same model (with a new classification head) was fine-tuned to HCD. } 
% \label{NLI_FT}
% \end{table}   

%% file: sections/2RelatedWork.tex
\subsection{Pre-trained Language Models}
From their inception when introduced in \cite{attentionIsAllYouNeed}, the use of the Transformer architecture has evolved greatly over the years. \cite{bert} released BERT, a transformer-based language model pre-trained on a 3.4 billion word corpus which produced state-of-the-art results on popular NLP benchmarks. Modern adaptations of the BERT pre-training paradigm such as RoBERTa \cite{roberta} and DeBERTa-v3 \cite{DeBERTa-v3} improve upon BERT by optimizing the pre-training tasks used to generate high-quality contextual token representations. We explore all three transformer models in this study. 

Additionally, we investigate the benefits of medical training data on HCD via Bio+Clinical BERT \cite{bio_clinical}, which is the BERT framework described above, but instead of being pre-trained on common texts, it is trained on biomedical corpora and clinical notes. Finally, we evaluate the performance of static sentence embeddings using SBERT \cite{reimers-2019-sentence-bert}. This experiment provides a highly regularized transformer baseline where only the classification head is fine-tuned for HCD.  

\subsection{Natural Language Inference}
The task of Natural Language Inference (NLI) is a multi-class pairwise classification problem for which, given a premise and a hypothesis, the model is tasked with predicting if the premise, Entails, Contradicts, or is Neutral towards the hypothesis. NLI data sets such as MNLI \cite{mnli_paper} are commonly used to evaluate a language model's ability to reason about complex language understanding tasks. NLI and HCD both depend on the compare and contrast of two disjoint texts. Additionally, it is reasonable for one to posit that there is a relationship between the notion of a pairwise contradiction and conflict. Thus in this study, we explore the relationship between notions of \textit{conflict} and \textit{contradiction} in effort to identify if NLI can be useful in the prediction of conflicting health information.

\subsection{Health Misinformation Detection}
Recently there has been increasing interest in misinformation detection for specific health topics from social media data \cite{sager2021identifying, elsherief2021characterizing, weinzierl2021misinformation}. ElSherief et al. focus on detecting misinformation regarding medications used for opioid use disorder (OUD) treatment from multiple social media using traditional machine learning models and BERT \cite{elsherief2021characterizing}. 
% They collect from four social media platforms, namely, Twitter, Youtube, Reddit and a drug forum. 
They formulate the problem as a binary classification task where a positive class refers to a post discussing a piece of misinformation (or myths) challenging OUD treatment and the negative class refers to any post that was not relevant. They report that logistic regression on TF-IDF based features perform better than BERT. Weinzierl et al. focus on identifying adoption or rejection of COVID-related misinformation on Twitter to better understand the effect of exposure to misinformation on COVID-19. They formulate the problem as stance classification and develop an ensemble architecture consisting of graph attention networks and BERT \cite{weinzierl2021misinformation}. Both HCD and health misinformation detection aim to measure information quality and improve public interactions with online health texts.

% The solution utilizes lexical, emotion, and semantic knowledge to classify the potential stance of tweets with respect to COVID-19 misinformation. 
% Like HCD, health misinformation is rooted in the analysis of health information quality via the characterization of health texts. 
% \cite{bert}. 
% In another relevant research, authors develop an intelligent bots to identify and respond to misinformation on Reddit regarding dermatology \cite{sager2021identifying}. 

%% file: sections/6Limitations_FutureWork.tex
\paragraph{Limitations:}

Although the human-in-the-loop approach using mTurk yielded reasonable conflicting health advice pairs, the resulting dataset has a couple of limitations. First we found that mTurk workers do not generate health conflict data which stylistically reflects real-world data. Specifically, mTurk workers struggle to write diverse sub-typical conflicts and often generated repetitive instances of conditional or temporal conflicts with no or limited stylistic and semantic variation. In general, synthetic samples have much more textual overlap than real samples, as mTurk annotators can often change/add a small number of tokens to generate a conflict. These themes are common and produce a stylistic drift between real and synthetic samples in our dataset. However, this is an expected limitation of mTurk-generated samples as they were only tasked with mimicking the linguistic phenomena of an advice conflict, not to generate realistic samples.\footnote{For more details on how Turkers were instructed to write each conflict type, please refer to the supplementary materials.}

Another limitation of this work is that the distribution of conflict-types found in this dataset will be biased towards the official data sources used to collect real advice statements. As the HCD task moves to new data domains such as social media data, different conflict types/distributions may emerge. 

The solution provided in this paper is limited to understanding the problem linguistically. In the real world, health conflicts are often subjective to user physiology, diagnosis and prognosis. Effects of health conflicts might differ in terms of degree of risk and potential severity. A user-centric conflict detection system should be personalized to avoid false alarms and cause user anxiety. This demands continuous user sensing, as information of the users' medical context is required for the personalized system to detect what is conflicting \textit{for a given user}. For example, the HCD system can only flag potentially dangerous drug-to-drug interactions if it knows what medications the end-user is currently taking. 

% Utilizing end-user medical information in this way might raise various ethical and privacy concerns and will require careful consideration in later iterations of this work. 

% Finally, a long-term goal of this project is to provide a personalized health watchdog app which can aid and accompany online health information consumers as they receive health advice. Health conflicts are often subjective to user physiology, diagnosis and prognosis. Effects of health conflicts might differ in terms of degree of risk and potential severity. So, a user-centric conflict detection system should be personalized to avoid false alarms and cause user anxiety and decision fatigue. This requires extensive continuous user sensing and monitoring, as information of the users' medical context is required for the personalized system to detect what is conflicting \textit{for a given user}. Utilizing end-user medical information in this way might raise various ethical and privacy concerns and will require careful consideration in later iterations of this work. 

% -Limitations: focusing on only English language... 
% in the future can be expanded on multiple languages using language independent meaning representations, such as AMR

\paragraph{Future work:}

Future work on HCD should investigate better use of medical knowledge. While experimental results on Bio+Clinical BERT showed HCD does not benefit from pre-training on scientific texts and clinical notes, it may be the case that HCD requires a medical knowledge integration solution that is well-suited for  non-clinical text related to public health. Medical knowledge graphs like UMLS \cite{bodenreider2004unified} may aid BERT-like models in the understanding of rare diseases and medications not common in language model vocabularies. 

Additionally, our attempt at organic HCD dataset expansion only targeted professional sources. Future works may consider non-professional sources such as social media data. However, advice extraction from social media data is itself a separate research question. Additionally, non-professional sources may provoke differing data distributions with new conflict types. Exploration of such data sources should be the subject of future works. 
% \begin{itemize}
%     \item [FW] knowledge integration
%     % \item [FW] include semantic meaning representation to BERT/ DeBERTa
%     % \item [FW] ensemble of different pre-trained language models as in \cite{li2021exploring} "[Exploring Text-transformers in AAAI 2021 Shared Task: COVID-19 Fake News Detection in English]: five model ensemble: BERT, Roberta, XL-net, Electra, Ernie"
    
% \end{itemize}

%% file: sections/7BroaderImpacts.tex
% Authors should take care to discuss both positive and negative outcomes. Authors are also expected to describe steps taken to prevent or mitigate potential negative outcomes

% Broader perspective, ethics and competing interests: In order to provide a balanced perspective, authors are required to include a statement about the potential broader impact of their work and ethical considerations. This statement needs to, at the bare minimum, be presented in a clearly marked paragraph/subsection/section in your paper. Authors should take care to discuss both positive and negative outcomes. Authors are also expected to describe steps taken to prevent or mitigate potential negative outcomes. For full papers that have collected new datasets and for dataset papers, discuss ethical considerations in the data collection process, and considerations around its release, and both potentially positive and negative outcomes of its use by others.

% +can be used to explain/ characterize online misinformation 

\section{Broader Impacts} 

\paragraph{New dataset with high quality and reliability: }
We provide a new public dataset on health conflict detection. The data is reliable as our use of multi-stage human-review helps mitigate annotation artifacts and human errors commonly found in large-scale crowd-sourced datasets. Additionally, our work on expanding the original HCD dataset uses data sourced from a wide variety of topics on various diseases and conditions. This makes HCD more health topic inclusive. 

% \textcolor{blue}{
% +new dataset
% +use multistage review process to mitigate potential annotator bias, annotation artifacts, and human errors: under-labeling, generating wrong samples, generating monotonous samples, label leakage corruption/memorization from the training to test data

% +inclusive about topics: refer topic distribution graph: diseases, conditions, lifestyle (food exercise). Extend the original preclude dataset to more diverse topics}
% -- New data is not focused on chronic disease. 

\paragraph{Providing consistent online health information:} An accurate health conflict detection system will protect consumers of online health information by providing a cautionary signal upon detection of potential health information conflicts. In addition, such technology can reduce the number of errors in medical decisions made by patients using online medical information sources. This is particularly important for individuals who rely on online sources for decision-making due to limited access to healthcare, an increasingly common case since the onset of a global pandemic. 

\paragraph{Reducing cognitive overload and aiding patient-provider communication:} In general, health information inconsistency checking is a challenging task for most humans. It can be overwhelming for patients to memorize all of the health advice they receive and cross-reference it with all the new pieces of advice they encounter. Automatic conflict detection can help patients identify potential conflicts and consider when to raise questions to their medical providers. This is especially important for caregivers of individuals with cognitive challenges such as dementia or those who suffer from extreme stress, where mental cross referencing between sources becomes increasingly difficult. Even for the general population, HCD reduces the burden to understand and memorize all of the fine-grained details of their medical conditions/medications by automatically detecting conflicts \textit{which pertain to a patient's specific condition.} Also, the goal of this work is to flag potential conflicts found in textual health information so an end-user can bring them to the attention of their medical provider. Thus there is no conflict resolution currently in-place, as it is beyond the scope of this work.

\textbf{Extending HCD task for detecting and characterizing misinformation:}
We additionally note that, while the non-augmented HCD samples in this study were taken from verified medical sources, the ideas proposed in this work extend to non-official sources such as text collected from Twitter, Reddit, Facebook or other social media \cite{sager2021identifying, elsherief2021characterizing, weinzierl2021misinformation}. For example, one can formulate the comparison of health advice between official social media sources (e.g. @CDCgov on twitter) and non-official sources aiming to spread health misinformation. This differs from other health misinformation formulations where the veracity of a claim is verified using the text itself, or an external scientific resource. In the HCD formulation, we not only better permit the direct comparison of two pieces of advice, but can predict fine-grained labels that provide better insight into \textit{how} two advice texts differ. Thus it can help to characterize misinformation from a linguistic standpoint. Additionally, fact checking often uses knowledge base triplets for claim verification. Under this paradigm, it may be hard to extend misinformation detection to new/niche health domains for which knowledge triplets are unavailable. Exploring misinformation detection in the context of HCD and language models may help keep predictive models more robust to low-resource domains or outdated evidence. 

% \textcolor{blue}{How HCD is related to misinformation? (i) enable to characterize misinformation, temporal, quantitative, conditional..thus extending fact checking, (ii) fact checking utilizes knowledge base triplets: which is not available for new domains, and can be outdated as new evidence emerge. We need more robust solution like PLMs and HCD explorations can generate insights. }

% \textbf{Applications to other tasks and domains:}
% Detecting conflicting health information can be helpful for deep semantic inference of other related medical NLP tasks, such as medical dialogue summarization and generation \cite{nobles2020examining, lin2021graph, sager2021identifying}, stance detection \cite{mutlu2020stance, weinzierl2021misinformation}, and fine-grained opinion mining \cite{gurukar2020towards, chen2021research}. 
% In addition, the concept of conflicting information can be extended to other domains including law, finance, and  politics. The findings from this paper can be helpful to design and develop the synthetic data augmentations for these additional domains. 
% % \todo[inline]{should talk about potential applications of this work for medical dialogue summarization, stance detection, opinion mining and distance in opinion. Also about different domains: law, finance, health, politics}

\section{Ethical Considerations}
This research involved human subjects only for data annotation, review, and prompt-based synthetic data generation. The project did not provide any intervention to any human subjects nor did it collect any user level data. So there is no risk to human subjects. The Amazon mechanical turk workers were recruited only for prompt-based synthetic data generation using formal procedure and no personal data was collected from them. The authors took careful measures to avoid annotation errors and maintain the quality and reliability of the data. Future progressions of HCD studies, in effort towards providing contextualized user-centric health information, may require use of end-user medical data. This might raise various ethical and privacy concerns and will require careful consideration. However, this is not a concern of the current study. 

The intended audience of the collected advice statements are often  for the general population. Thus, the HCD dataset may not contain many advice statements directed at underrepresented communities/populations. Future rounds of data collection should focus on both diversification of medical condition coverage and  target audience. User-generated medical content from social media can be a good source to collect health advice targeted to a specific sub-population, e.g. \textit{heart disease} related advice targeted to African Americans.  

%% file: sections/8Conclusion.tex
Detecting conflicting health information makes human interaction with online health platforms safer. In this work, we introduce the HCD task in the context of pre-trained language models. We provide a detailed analysis of five different language models, examining the challenges brought about when predicting different health conflict types. Our experimental results find that the DeBERTa-v3 architecture performs better on challenging conflict types with less obvious semantic patterns.  However, for simpler conflicts like conditional and temporal, BERT and RoBERTa provides comparable performance.  We additionally confirm that the token attribution scores provided by a deep interpretability model, Captum, align with human judgement regarding input importance for predicting different conflict types. Finally, we show that the addition of synthetic conflict data does help with the prediction of real-world data, irregardless of differences in style and content accuracy.